\documentclass[]{TEAI}

\usepackage{amsmath} 
\usepackage{natbib}
\usepackage{graphicx}
\usepackage{subcaption} 

\usepackage[toc,page,header]{appendix}
\usepackage[utf8]{inputenc} 
\usepackage[T1]{fontenc}    

\usepackage{url}            
\usepackage{booktabs}       
\usepackage{amsfonts}       
\usepackage{nicefrac}       
\usepackage{microtype}      
\usepackage{wrapfig}

\usepackage{amssymb}  
\usepackage{fontawesome}  
\usepackage{url}  

\usepackage{titletoc}

\usepackage{tikz}  
\usepackage{comment}  
\usepackage{tabularx}  
\usepackage{booktabs}  

\usepackage{minitoc}

\usepackage{booktabs}
\usepackage{array}
\usepackage{etoolbox}

\definecolor{lightblue}{RGB}{200, 230, 255}  
\definecolor{headerblue}{RGB}{150, 200, 255} 

\usepackage{pgfplots}
\usepackage[utf8]{inputenc} 
\usepackage[T1]{fontenc}    

\usepackage{url}            
\usepackage{booktabs}       
\usepackage{amsfonts}       
\usepackage{nicefrac}       
\usepackage{microtype}      
\usepackage{xcolor}         
\usepackage{graphicx}
\usepackage{float}
\usepackage{comment}
\usepackage{multirow} 
\usepackage{amsmath} 
\usepackage{makecell} 
\usepackage{siunitx}  
\usepackage{tikz}
\usepackage{pgf-pie} 
\usepackage{subcaption}
\usepackage{wrapfig}
\usepackage[export]{adjustbox}

\usepackage{ragged2e}      
\usepackage{tabularx}       
\usepackage{array}          
\usepackage{caption}        
\usepackage{enumitem}
\usepackage{pifont}
\usepackage[hang,flushmargin]{footmisc} 

\usepackage{tcolorbox}

\usepackage{tcolorbox}    
\tcbuselibrary{breakable}  
\tcbuselibrary{skins}      

\usepackage{tabularx}
\usepackage{listings}

\usepackage{times}
\usepackage{latexsym}

\usepackage[T1]{fontenc}

\usepackage[utf8]{inputenc}

\usepackage{microtype}

\usepackage{inconsolata}

\usepackage{graphicx}

\usepackage{amsmath}
\usepackage{amsfonts}

\usepackage[most]{tcolorbox}
\usepackage{listings}
\usepackage{ragged2e}
\usepackage{enumitem}
\usepackage{booktabs}
\usepackage[most]{tcolorbox}
\usepackage{listings}
\usepackage{ragged2e}
\tcbuselibrary{breakable,listings}

\usepackage{algorithm}
\usepackage{algorithmic}
\setboolean{ALC@noend}{true}
\usepackage{cuted} 

\definecolor{codegreen}{rgb}{0,0.6,0}
\definecolor{codegray}{rgb}{0.5,0.5,0.5}
\definecolor{codepink}{RGB}{252, 142, 172}
\definecolor{codepurple}{rgb}{0.58,0,0.82}
\definecolor{backcolour}{RGB}{245,245,245}
\lstdefinestyle{scaler}{
    backgroundcolor=\color{backcolour},   
    commentstyle=\color{magenta},
    keywordstyle=\color{blue},
    numberstyle=\tiny\color{codegray},
    stringstyle=\color{codepurple},
    basicstyle=\fontfamily{\ttdefault}\footnotesize,
    breakatwhitespace=false,        
    breaklines=true,                
    keepspaces=true,    
    frame=single,
    numbersep=5pt,                  
    showspaces=false,              
    showstringspaces=false,
    showtabs=false,               
    tabsize=2,
    classoffset=1, 
    keywordstyle=\color{violet},
    classoffset=0,
}
\newtcblisting{PromptBox}[1]{%
enhanced,
  breakable,
  listing only,
  colback=white,
  colframe=black!60,
  boxrule=0.8pt,
  arc=2pt,
  left=6pt,right=6pt,top=6pt,bottom=6pt,
  colbacktitle=black!60,
  coltitle=white,
  fonttitle=\bfseries,
  title={#1},
  before upper=\RaggedRight,
  segmentation style={draw=none},
  listing engine=listings,
  listing options={style=prompt}
}
\usepackage{amsmath}
\usepackage{array}
\usepackage{longtable} 

\usepackage{amsmath}
\usepackage{amssymb}
\usepackage{bm}
\usepackage{enumerate}
\usepackage{algorithm}
\usepackage{algorithmic}
\usepackage{longtable}
\usepackage[most]{tcolorbox}

\usepackage{enumitem}
\usepackage{multirow}
\usepackage{booktabs}

\newtcolorbox[
  auto counter
]{mybox}[1][]{%
  colback=blue!5!white,
  colframe=blue!75!black,
  arc=4pt,
  boxrule=0.6pt,
  left=6pt, right=6pt, top=4pt, bottom=4pt,
  title={},
  #1
}

\title{DenoiseRL: Bootstrapping Reasoning Models to Recover from Noisy Prefixes}

\author{
 \textbf{Caijun Xu\textsuperscript{1,2}},
 \textbf{Changyi Xiao\textsuperscript{1}},
 \textbf{Zhongyuan Peng\textsuperscript{1}},
 \textbf{Yixin Cao\textsuperscript{1,2,$\dagger$}},
}
\affiliation[1]{\mbox{Fudan University}} 
\affiliation[2]{\mbox{Shanghai Innovation Institute}}

\abstract{
\begin{abstract}

Reinforcement learning has become a central paradigm for advancing reasoning in large language models, yet most existing methods still depend on stronger teacher models or heavily curated difficult datasets, limiting scalable capability improvement. 
In this paper, we introduce DenoiseRL, a reinforcement learning framework that substitutes external supervision with recovery-oriented optimization over failures from weak models. 
Instead of relying on stronger supervision or carefully engineered data, DenoiseRL learns directly from noisy reasoning prefixes by converting them into opportunities for improvement, while exercising fine-grained control over the noise intensity, making training more scalable and effective. 
This yields a richer and more diverse learning signal, improving exploration efficiency by leveraging imperfect model behavior. 
Empirically, DenoiseRL consistently outperforms strong on-policy RL baselines across competitive mathematical reasoning tasks and interactive decision-making tasks, while promoting stronger self-corrective behavior as training difficulty increases, highlighting an effective and scalable pathway for improving agentic reasoning capabilities of large language models.
\end{abstract}
}

\correspondence{\email{cjxu25@m.fudan.edu.cn}}
\checkdata[Github]{\url{https://github.com/ALEX-nlp/DenoiseRL}}

\begin{document}
\maketitle
\renewcommand{\thefootnote}{}
\footnotetext{$^\dagger$Corresponding authors.}
\renewcommand{\thefootnote}{\arabic{footnote}}


\vspace{-1.5em}

\section{Introduction}

Reinforcement learning (RL) has emerged as a central post-training paradigm for large language models (LLMs), driving substantial advances in complex reasoning tasks~\cite{shao2024deepseekmath,hu2025reinforce++,xiao2025bnpobetanormalizationpolicy,ahmadian2024basicsrevisitingreinforcestyle,ouyang2022training,guo2025deepseek}. 
Despite these successes, state-of-the-art systems often rely on supervision or guidance from even stronger models~\cite{zhang2026reinforcementawareknowledgedistillationllm,yu2026knowrlboostingllmreasoning,he2025deepmath103klargescalechallengingdecontaminated}. 
This dependence exposes a structural limitation: when no sufficiently capable off-the-shelf teacher is available, further capability gains become increasingly difficult, raising a fundamental question: \emph{how can strong models be obtained without relying on stronger models as supervisors?}

To address this challenge, prior work has explored two main directions. The first is the weak-to-strong paradigm, which improves stronger models using supervision derived from weaker ones~\cite{liu-etal-2025-atlantis,geng2026weak}. 
While effective in practice, its performance is fundamentally constrained by the quality of the teacher signal and easily leads to noise in the training process~\cite{yuan2026incentivizingstrongreasoningweak,lang2025selectiveweaktostronggeneralization,yao2025revisiting}. 
The second direction focuses on increasing task difficulty through data construction~\cite{zhan2026mathsmithextremelyhardmathematical,liu2025synlogicsynthesizingverifiablereasoning} or curriculum design~\cite{xu2026scaler,parashar2026curriculumreinforcementlearningeasy}. 
However, these approaches typically depend on carefully engineered pipelines, complex filtering and verification procedures, and substantial human effort in data design and curation.

In this work, we propose \textbf{DenoiseRL}, a new RL paradigm that unifies weak-to-strong learning with difficulty-driven data synthesis. 
Instead of using weak models to synthesize hard data or provide learning signals, we repurpose weak models as generators of structured perturbations, which automatically increases training difficulty without generating new questions.
With a fine-grained control strategy of noise, DenoiseRL also implements a sample-level curriculum, continuously tracking boundaries of model capabilities and automatically constructing well-aligned training instances.
This enables scalable improvement of reasoning capability without relying on external teacher's trajectories and manually curated supervision.
It also casts reasoning RL as a denoising problem: weak-model errors serve as structured corruptions of the trajectory, and the policy learns to reconstruct a valid solution path from these corrupted states, echoing the principle of denoising autoencoders and BART-style pretraining~\citep{vincent2008extracting,lewis2020bart}.

\begin{figure*}[t]
    \centering
    \includegraphics[width=\columnwidth]{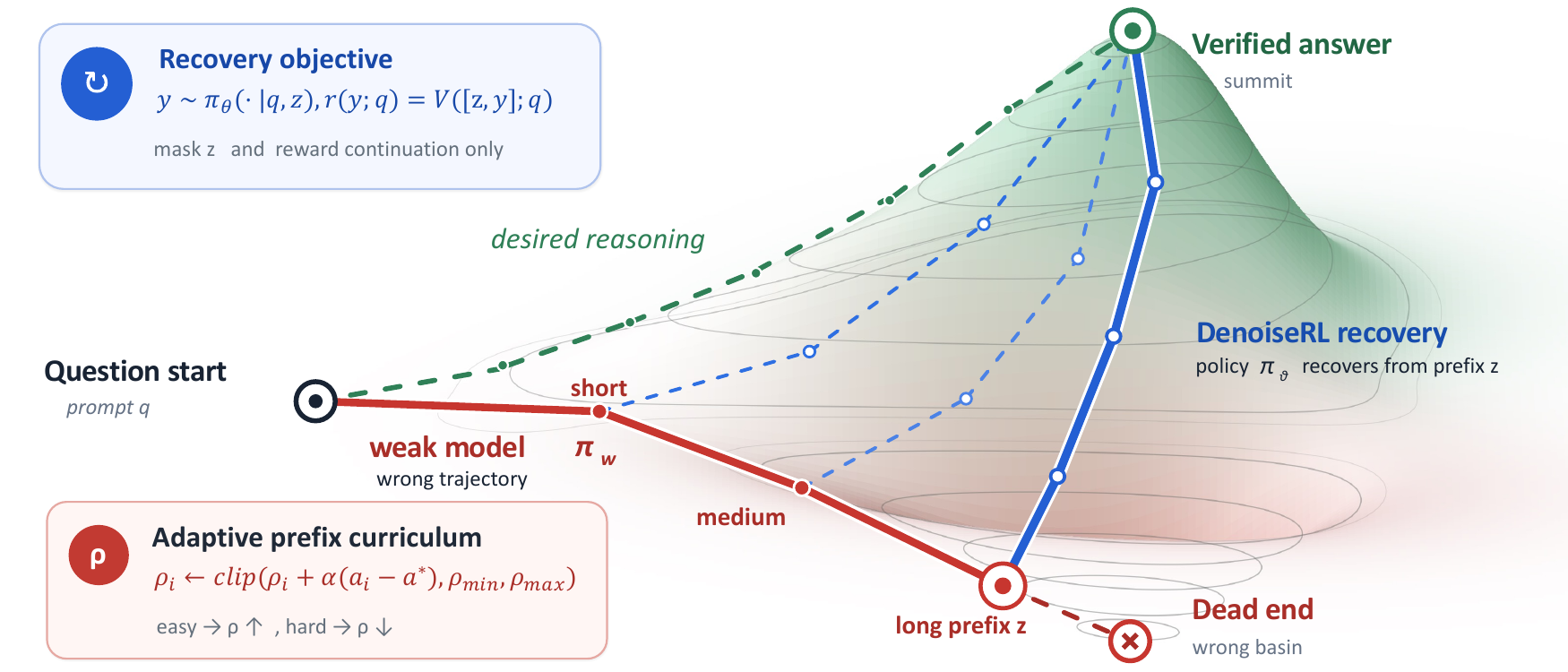}
    \caption{Overview of \textbf{DenoiseRL}.
    Starting from a prompt $q$, a weak policy $\pi_w$ generates an incorrect trajectory toward a dead end.
    DenoiseRL truncates this trajectory into a noisy prefix $z$ and adaptively adjusts its length through a sample-level curriculum: longer prefixes increase recovery difficulty for easy samples, whereas shorter prefixes reduce it for hard samples.
    Conditioned on $(q,z)$, the policy $\pi_\theta$ is trained to escape the wrong reasoning basin and recover toward a verified answer; prefix tokens are masked from the RL objective, and rewards are assigned only to verifier-approved completions.}
    \label{fig:main}
\end{figure*}

Specifically, as illustrated in Figure~\ref{fig:main}, we sample erroneous reasoning trajectories from weak models and inject their truncated prefixes into the policy's rollouts~\cite{setlur2026reuse}.
Rather than fixing the prefix length, DenoiseRL maintains a separate prefix ratio for each problem and adapts it online using the policy's recovery accuracy: the ratio increases when recovery is easier than a target level and decreases when recovery is too difficult.
This sample-level curriculum keeps the injected noise near the policy's evolving capability boundary, providing informative training signals throughout training.
The policy is then optimized to denoise these corrupted prefixes and complete reasoning trajectories correctly.

We choose the prefix as the injection point because it plays a disproportionate role in shaping the subsequent reasoning trajectory.
Even if the reasoning prefix—which ultimately leads to a dead end—is correct in the short term, it still disrupts the model's reasoning space and is prone to causing reasoning failure. 
Prior work shows that high-quality prefixes can steer the policy toward more favorable reasoning states and improve RL efficiency through prefix-level conditioning or optimization~\cite{chen2025llmssignaltheyreright,sun2026well}. 
DenoiseRL reverses this perspective: we inject erroneous weak-model prefixes as structured noise, thereby controlling the starting state of reasoning and forcing the policy to recover from corrupted intermediate states.
This mechanism induces two tightly coupled effects. First, it substantially expands the diversity of training states, since noisy prefixes span a much broader space of failure modes than correct trajectories, exposing the policy to off-policy contexts that are rarely encountered in standard on-policy RL~\cite{kim2025training,cai2025training,yan2026learning}. Second, it directly strengthens a critical yet underdeveloped capability: recovery from mistakes. 
Rather than continuing along incorrect intermediate conclusions, the model is required to explicitly revise and correct its reasoning. 
By embedding erroneous prefixes into the optimization objective, DenoiseRL elevates self-correction from an emergent behavior to a direct training target~\cite{huang2024large,welleck2022generating}.



We summarize our contributions as follows:
\begin{itemize}
    \item We propose DenoiseRL, a recovery-oriented RL framework that repurposes weak-model failures as structured reasoning-state noise, enabling improvement without stronger teachers or manually curated hard data.
    \item We introduce a sample-level curriculum that adapts each problem's noise intensity from online accuracy, keeping training near the policy's evolving capability boundary.
    \item We show consistent gains over on-policy  baselines across mathematical reasoning and agentic tasks, and identify external noise and fine-grained control as key design choices.
\end{itemize}

\section{Related Work}

\paragraph{Bootstrapping Reasoning via On-Policy RL.}
Outcome- and process-driven RL have superseded supervised fine-tuning for scaling reasoning capabilities~\citep{ouyang2022training, khalifa2025process,guo2025deepseek}.
Frameworks such as GRPO~\citep{shao2024deepseekmath} and DAPO~\citep{yu2026dapo} drive this progress, yet they are fundamentally bounded by the model's self-generated state distribution.
Once the policy saturates, it predominantly generates correct rollouts or narrowly confined failure modes, creating an exploration bottleneck where informative failures are too scarce for meaningful gradient updates~\citep{kim2025training, cai2025training,liu2025evocot}.

\paragraph{Weak-to-Strong Generalization (W2SG).}
To break capability plateaus, W2SG~\citep{burns2023weak,somerstep2024statistical,geng2026weak} leverages weaker models to supervise highly capable students.
However, this paradigm inherently caps the student's ceiling: the strong policy is optimized to imitate pseudo-labels, making it vulnerable to the weak supervisor's noise and limited capacity~\citep{yao2025revisiting}.
Rather than treating the weak model as an imperfect oracle, DenoiseRL inverts its role, utilizing it strictly as a low-cost generator of out-of-distribution mistakes.


\paragraph{Prefix-Conditioned and Off-Policy Exploration.}
Another line of work improves exploration by injecting external prefixes or off-policy trajectories into RL.
LUFFY~\citep{yan2026learning} mixes off-policy reasoning traces with on-policy RL, while PrefixRL~\citep{setlur2026reuse} conditions on successful off-policy prefixes and optimizes the remaining continuation.
More broadly, prefix- and trajectory-guided methods use expert solutions, oracle hints or successful traces to make sparse-reward problems more reachable~\citep{qu2026pope,sun2026well,kim2025training}.
DenoiseRL differs by using weak-model prefixes not as privileged hints, but as misleading reasoning states from which the policy must recover.
\section{Method}

We propose DenoiseRL, a denoising reasoning framework that trains LLMs to recover from noisy reasoning spaces.
Turning this idea into a training algorithm raises two questions: what the policy is asked to denoise, and how strong the injected noise should be as training proceeds.
Section~\textit{Denoising Reasoning with Noisy Prefixes} formalizes prefixes of weak-model failures as structured reasoning-state noise and defines the corresponding training objective.
Section~\textit{Adaptive Denoise Curriculum Learning} then controls the noise level for each problem using the policy's recovery accuracy and refreshes problems whose difficulty has stabilized.

\subsection{Denoising Reasoning with Noisy Prefixes}
\label{section:3.1}

Denoising is a long-standing way of turning corruption into supervision~\citep{vincent2008extracting,lewis2020bart}, while this corruption requires no annotation and makes the learned solution robust to perturbation.
Reasoning RL needs precisely these properties: when the verifier remains the same, being \textit{noisier} implies greater difficulty; a policy model also requires this kind of robustness against interference~\cite{shi2023largelanguagemodelseasily,anantheswaran2025cuttingnoiseboostingllm}.

As perturbing the problem statement changes the task and usually invalidates its ground-truth answer, we move the corruption from the input to the reasoning space: the problem and its answer are left untouched, while the intermediate state the policy reasons from is perturbed.

\paragraph{Defining noise in the reasoning space.}
Let $\pi_\theta$ be the policy we optimize over a pool of training problems $\mathcal{D}$, and let $\mathcal{V}(y;q)\in\{0,1\}$ be the rule-based verifier that scores a response $y$ to a problem $q\in\mathcal{D}$ by comparing its final answer with the ground truth $a^\star(q)$.
Any token sequence $z$ may be placed at the beginning of the response, after which the policy continues as $\pi_\theta(\cdot\mid q,z)$. 
We measure the intensity of $z$ by the gap between the expected verifier rewards of clean and prefix-conditioned generation:
\begin{equation}
\begin{aligned}
\mathrm{Intensity}(z;q)
:= \mathbb{E}_{y\sim\pi_\theta(\cdot\mid q)}
\big[\mathcal{V}(y;q)\big]
- \mathbb{E}_{y\sim\pi_\theta(\cdot\mid q,z)}
\big[\mathcal{V}([z,y];q)\big].
\end{aligned}
\label{eq:noise-def}
\end{equation}
Any $z$ with $\mathrm{Intensity}(z;q)>0$ can be regarded as noise in the reasoning space of problem $q$.

\paragraph{Prefixes of wrong trajectories as noise.}
The beginning of a trajectory is the most consequential place to intervene: a well-begun prefix steers the policy toward favorable reasoning states, while a model that has committed to a flawed direction rarely backtracks~\citep{huang2024large,lei2026stepbackprefiximportance}.
DenoiseRL exploits the same leverage in reverse, taking as noise $z$ a prefix of an incorrect solution produced by a weak model $\pi_{\mathrm{w}}$, which is substantially less capable than $\pi_\theta$.
Such a prefix need not contain any visibly incorrect step, yet it already carries the commitment that leads to the wrong final answer, so empirically continuing from it decreases the expected verifier reward under prefix-conditioned generation and yields $\mathrm{Intensity}(z)>0$, which makes it valid noise.

\paragraph{Denoising as the training objective.}
DenoiseRL makes escaping such noisy states the training objective: rollouts begin from a noisy prefix, and the policy still receives the terminal reward $r(y;q)=\mathcal{V}(y;q)$ for reaching the verified answer from there.
As the prefix is not sampled by the current policy, we assign its tokens a zero loss mask and update only the on-policy continuation. 
While recovery capability in standard RL is only  treated as a by-product, DenoiseRL turns it into a clear, trainable objective.

Then difficulty  becomes an adjustable variable, since revealing more of wrong solutions hardens the same problems.

\subsection{Adaptive Denoise Curriculum Learning}
\label{section:3.2}
To control task difficulty and maintain effective training signals throughout training, we can view adjusting the length of noisy prefixes as a means of controlling the intensity of noise.
Following~\cite{xu2026scaler}, DenoiseRL maintains a separate prefix ratio for each problem to achieve fine-grained control and uses observed recovery accuracy as online feedback.
Once a problem's ratio has stabilized, the problem is replaced to preserve a long-term learning signal.

\paragraph{Accuracy-driven ratio updates.}
At training step $s$, we maintain an active set $\mathcal{A}_s$ of $B$ problems.
Each problem $q_i\in\mathcal{A}_s$ carries a ratio $\rho_i^{(s)}\in[\rho_{\min},\rho_{\max}]$, initialized at $\rho_{\mathrm{init}}$, that specifies the desired fraction of its failed trajectory $w_i$ to reveal.
All $K$ rollouts for $q_i$ use the same current ratio.
Let $z_i^{(s)}$ be the noise of problem $q_i$ at step $s$, $c_{i,k}\in\{0,1\}$ be the verifier correctness of rollout $y_{i,k}\sim\pi_\theta(\cdot\mid q_i,z_i^{(s)})$, and
\begin{equation}
a_i^{(s)}
=
\frac{1}{K}\sum_{k=1}^{K}c_{i,k}
\label{eq:recovery-accuracy}
\end{equation}
be the resulting per-problem recovery accuracy.
After collecting these rollouts, we update the ratio toward a target accuracy $a_{\mathrm{target}}$:
\begin{equation}
\rho_i^{(s+1)}
=
\mathrm{clip}\!\left(
\rho_i^{(s)}
+\alpha\big(a_i^{(s)}-a_{\mathrm{target}}\big),
\rho_{\min},
\rho_{\max}
\right),
\label{eq:dynamic-noise-update}
\end{equation}
where $\alpha>0$ is the adaptation step size.
If the problem with current noise is easier than the target, the ratio increases and exposes more of the failed trajectory; if the problem is  too difficult, the ratio decreases.
The controller therefore tracks the noise level at which each problem remains informative as the policy improves.

\paragraph{Line-Aligned Noisy Prefixes.}
To ensure that the injected noise represents a coherent but potentially flawed reasoning state rather than an arbitrary syntactic truncation, we round the requested prefix length to the nearest complete line.
Write $w_i$ as $L_i$ lines, let $t_{i,\ell}$ be the token count through the end of line $\ell$, and define the set of admissible boundaries
$\mathcal{B}_i=\{0,t_{i,1},\ldots,t_{i,L_i}\}$, where $t_{i,L_i}=|w_i|$.
The continuous ratio used at step $s$ is realized as
\begin{equation}
\gamma_i^{(s)}
=
\operatorname*{arg\,min}_{t\in\mathcal{B}_i}
\left|t-\rho_i^{(s)}|w_i|\right|.
\label{eq:dynamic-prefix-length}
\end{equation}
Thus, $\gamma_i^{(s)}$ is the line boundary nearest to the nominal token budget $\rho_i^{(s)}|w_i|$.
For a single-line response with no internal line boundary, we instead use the nearest token position, as $\gamma_i^{(s)} = \rho_i^{(s)}|w_i|$.
Therefore, the noise $z_i^{(s)} = w_{1:\gamma_i^{(s)}}$.

\paragraph{Curriculum refreshing.}
To sustain diverse training signals, we periodically replace problems whose prefix ratios have stabilized. For each problem, we estimate the recent trend of its prefix ratio using a least-squares slope over the last $H$ updates.
We index them  as
$r_{i,h}=\rho_i^{(s-H+h)}$ for $h=1,\ldots,H$.
The slope of their least-squares linear fit against the update index is
\begin{equation}
\beta_i
=
\frac{
\sum_{h=1}^{H}(h-\bar{h})(r_{i,h}-\bar{r}_i)
}{
\sum_{h=1}^{H}(h-\bar{h})^2
},
\label{eq:curriculum-least-squares}
\end{equation}
where $\bar{h}=(H+1)/2$ and
$\bar{r}_i=\frac{1}{H}\sum_{h=1}^{H}r_{i,h}$.
Thus, $\beta_i$ estimates the per-step trend of the recent prefix ratios.
After at least $H$ observations, we regard a problem as stable when
\begin{equation}
|\beta_i|\leq\tau,
\label{eq:curriculum-stability}
\end{equation}
where $\tau$ is a small stability threshold.
Stable problems are retired and replaced by unseen problems from the current pass through $\mathcal{D}$.
To place a new problem near the active set's current difficulty, it inherits the post-update mean ratio, computed before replacement:
\begin{equation}
\rho_{\mathrm{new}}
=
\frac{1}{|\mathcal{A}_s|}
\sum_{q_j\in\mathcal{A}_s}
\rho_j^{(s+1)}.
\label{eq:replacement-rho}
\end{equation}
As DenoiseRL changes only rollout construction and masks, it can be combined with other RL backbones.

\section{Experiments}


\begin{table*}[t]
\centering
\caption{
Main results on mathematical and reasoning benchmarks.
For AMC23, AIME2024, and AIME2025, we report AVG@16; for MATH500 and BBEH, we report AVG@1.
Results are grouped by base model. Within each group, the best result is shown in \textbf{bold} and the second-best result is \underline{underlined}. 
Tied second-best results are all underlined.
}
\label{table:1}

\small
\setlength{\tabcolsep}{4.5pt}
\renewcommand{\arraystretch}{1.08}

\begin{tabular*}{0.92\textwidth}{@{\extracolsep{\fill}}lcccccc@{}}
\toprule
\textbf{Method} 
& \textbf{MATH500} 
& \textbf{AMC23} 
& \textbf{AIME24} 
& \textbf{AIME25} 
& \textbf{BBEH} 
& \textbf{Avg.} \\
\midrule

\multicolumn{7}{c}{\textit{Qwen3-4B-Base}} \\
\midrule
Base 
& 70.0 & 43.1 & 8.3 & 7.7 & 4.1 & 26.6 \\
GRPO 
& 83.6 & 63.1 & 22.1 & 18.1 & 11.1 & 39.6 \\
DAPO 
& 83.8 & 62.5 & 20.6 & 21.5 & 10.4 & 39.8 \\

Critique-GRPO (Critique w/ GT) 
& \textbf{86.2} & 61.6 & 22.5 & 21.3 & 11.1 & 40.5 \\

GRPO + Correct Prefix 
& 80.2 & 58.9 & 18.5 & 12.3 & \underline{13.3} & 36.6 \\

\textbf{DenoiseRL-GRPO} 
& \underline{85.6} & \textbf{68.4} & \textbf{24.8} & \underline{21.7} & 10.7 & \underline{42.2} \\
\textbf{DenoiseRL-DAPO} 
& 83.6 & \underline{67.2} & \underline{23.3} & \textbf{22.9} & \textbf{14.3} & \textbf{42.3} \\

\midrule
\multicolumn{7}{c}{\textit{Qwen3-8B-Base}} \\
\midrule
Base 
& 70.4 & 49.2 & 11.9 & 10.8 & 4.1 & 29.3 \\
GRPO 
& \underline{87.8} & 69.7 & 24.0 & 22.9 & 10.6 & 43.0 \\
DAPO 
& 87.0 & 69.7 & 23.8 & 21.7 & 11.7 & 42.8 \\

Critique-GRPO (Critique w/ GT)  
& 86.6 & 68.8 & 26.0 & 22.1 & \underline{13.5} & 43.4 \\

GRPO + Correct Prefix 
& 85.6 & 63.6 & 19.8 & 16.3 & 12.8 & 39.6 \\

\textbf{DenoiseRL-GRPO} 
& \textbf{88.0} & \underline{71.4} & \underline{26.3} & \textbf{23.8} & \textbf{15.4} & \underline{44.9} \\
\textbf{DenoiseRL-DAPO} 
& \underline{87.8} & \textbf{73.1} & \textbf{28.8} & \underline{23.3} & 13.0 & \textbf{45.2} \\

\bottomrule
\end{tabular*}
\end{table*}

\subsection{Settings}
\paragraph{Noisy Prefix Collection.}
We use Qwen2.5-1.5B-Instruct~\cite{qwen2} as the weak model to collect noisy reasoning trajectories. We sample 8 responses for each problem in MATH-7.5K~\cite{hendrycks2021measuring} and retain problems with a valid incorrect boxed answer.

\paragraph{Reinforcement learning.}
We train Qwen3-4B-Base and Qwen3-8B-Base~\cite{qwen3technicalreport} as policy models on MATH-7.5K. DenoiseRL uses an active batch of $B=16$ problems and samples $K=16$ rollouts per problem. 
The $K$ rollouts of a problem share its current prefix ratio. 
We initialize $\rho_{\mathrm{init}}=\rho_{\min}=0$ and use $\rho_{\max}=0.5,a_{\mathrm{target}}=0.75$ for both models. 
The per-problem curriculum uses step size $\alpha=0.2$, a history window of $H=5$ and stability threshold $\tau=0.02$. 
Stable problems are replaced by new ones, which inherit the post-update mean prefix ratio of the active batch.

We use a maximum response length of $R=4096$ tokens. 
For all runs, we use a learning rate of $10^{-6}$, no KL regularization, and PPO~\cite{schulman2017proximal} clipping bounds $\varepsilon_{\mathrm{low}}=\varepsilon_{\mathrm{high}}=0.2$. Training rollouts use temperature $1.0$ and top-p $1.0$. 
We keep the same hyperparameters for all experiments, and provide the training prompt and more training hyperparameters  in Appendix~A of the supplementary material, while response-length and training efficiency analyses are discussed in Appendix~B.
All training and evaluation experiments were conducted on  4  NVIDIA H100 GPUs.

\paragraph{Evaluation.}
We evaluate models on reasoning benchmarks including MATH500~\cite{lightman2024let}, AMC23~\cite{aops_amc}, AIME2024, AIME2025~\cite{aops_aime}, and BBEH~\cite{suzgun2023challenging}. Validation decoding uses temperature $=0.6$, top-p $=0.95$.
We report AVG@16 for AIME2024, AIME2025, and AMC23, and AVG@1 for the remaining benchmarks.

\paragraph{Baselines.}
We compare our denoise training against the base model and several RL baselines, GRPO~\cite{shao2024deepseekmath}, DAPO~\cite{yu2026dapo}, a self-correct baseline Critique-GRPO~\cite{zhang2025critique} and GRPO with correct prefixes from Qwen2.5-1.5B-Instruct using the same pipeline of noisy prefix collection.

\subsection{Mathematical Reasoning Performance}

Table~\ref{table:1} summarizes the main reasoning results.
Without a teacher model stronger than the policy, DenoiseRL uses only incorrect prefixes from a smaller model yet achieves the best overall performance.
On Qwen3-4B-Base, DenoiseRL-GRPO and DenoiseRL-DAPO improve over their corresponding baselines by 2.6 and 2.5 average points, respectively; on Qwen3-8B-Base, the gains are 1.9 and 2.4 points.
DenoiseRL-DAPO achieves the highest average at both scales, reaching 42.3 and 45.2.
The improvements are also broad across benchmarks: the two DenoiseRL variants jointly achieve the best result in five of the six reported columns for the 4B model and in every column for the 8B model.
In contrast, adding correct prefixes alone underperforms standard GRPO, suggesting that the gains come from learning to recover from errors rather than from prefix conditioning.

Beyond improving accuracy, denoising directly trains the policy to recover from flawed reasoning states, thereby strengthening self-correction.
Both DenoiseRL variants outperform Critique-GRPO, a self-correction baseline, and the best DenoiseRL result exceeds it by 1.8 points at each model scale.
This comparison shows that recovery training over structured errors provides a more effective self-correction signal than critique-based training in our setting.

\subsection{Effect of Noisy Prefix Source}
\label{sec:noise-source}
To investigate how different noisy-prefix sources affect downstream performance and the noise intensity defined in Equation~\eqref{eq:noise-def}, we compare three sources: incorrect prefixes generated by the frozen pre-RL policy, random-token prefixes, and incorrect prefixes generated by the external weak model used by DenoiseRL.

\paragraph{Impact on performance.}
Under dynamic noise control, we allow random-token prefixes to contain up to $1{,}024$ noise tokens, while the pre-RL self and small-model sources use the same dynamic prefix-ratio configuration as the main DenoiseRL-GRPO experiment.
Table~\ref{tab:noise-source-performance} reports the average over the four mathematical benchmarks, BBEH accuracy, and the overall average over all five benchmarks.

\begin{table}[t]
    \centering
    \caption{Performance of different noisy-prefix sources under dynamic noise control (\%).}
    \label{tab:noise-source-performance}
    \small
    \setlength{\tabcolsep}{3.5pt}
    \begin{tabular*}{\columnwidth}{@{\extracolsep{\fill}}lccc@{}}
        \toprule
        \textbf{Method} & \textbf{Math Avg.} & \textbf{BBEH} & \textbf{Avg.} \\
        \midrule
        Random tokens & 48.2 & 9.8 & 40.5 \\
        Pre-RL self & 47.9 & \textbf{12.2} & 40.7 \\
        \textbf{DenoiseRL-GRPO} & \textbf{50.1} & 10.7 & \textbf{42.2} \\
        \bottomrule
    \end{tabular*}
\end{table}

DenoiseRL-GRPO, which uses prefixes generated by an external small model, achieves the best mathematical and overall averages.
These results suggest that structured noise introduced by an external model is more effective than either unstructured random tokens or errors sampled from the policy's own pre-RL checkpoint.

\paragraph{Online estimation of noise intensity.}
To further investigate the impact of different noise sources on noise intensity, we fix the prefix ratio at $0.2$ for pre-RL self and small model noise, 256 tokens for random token noise, sampling 12 standard GRPO rollouts and 4 denoise rollouts for each problem.

\begin{figure}[t]
    \centering
    \includegraphics[width=0.5\columnwidth]{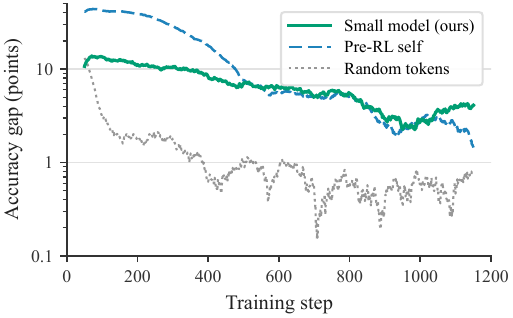}
    \caption{Online estimates of noise intensity for three prefix sources through step $1{,}200$. Intensity is the standard-rollout accuracy minus the noisy-rollout accuracy at each training step; curves show a 100-step moving average on a logarithmic vertical scale.}
    \label{fig:noise-source-intensity}
\end{figure}

Figure~\ref{fig:noise-source-intensity} estimates the noise intensity  using the accuracy gap between standard and noisy-prefix rollouts, where a larger positive gap indicates higher intensity.
Pre-RL self prefixes have the highest initial and overall intensity ($14.9$ points) but decay sharply, while small-model prefixes are weaker on average ($6.9$) yet more persistent; random tokens are the weakest ($1.9$).
During steps $1{,}000$--$1{,}200$, small-model prefixes retain the highest mean intensity ($3.9$ versus $2.2$ for pre-RL self and $0.6$ for random tokens), providing the strongest late-stage noise.
These results show that useful training noise is characterized not only by high initial intensity, but also by non-trivial intensity after the policy has learned to recover: both model-generated sources satisfy this criterion, while small-model prefixes, the external structured noise, provide the strongest late-stage intensity.

\subsection{Adaptive Control of Noise Intensity}
\label{sec:adaptive-noise-control}

To test whether controlling noise intensity is important beyond simply injecting a fixed amount of noisy reasoning, we conduct a controlled comparison on Qwen3-4B-Base. 
We consider three strategies under the same denoising-RL backbone and evaluation protocol: a fixed prefix ratio of $0.2$, a coarse-grained controller that adjusts a shared noise ratio using aggregate training accuracy, and our fine-grained controller that maintains and updates a separate ratio for each problem using its recovery accuracy.

\begin{figure}[t]
\centering
\includegraphics[width=0.5\linewidth]{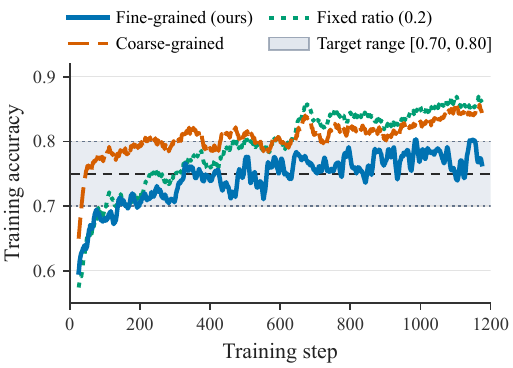}
\caption{
Target tracking under different noise-intensity control strategies on Qwen3-4B-Base. 
Curves show the 50-step moving average of training recovery accuracy over the first 1,200 updates. The shaded region denotes the target
interval $a_{\mathrm{target}}\pm0.05=[0.70,0.80]$.
}
\label{fig:adaptive-control-tracking}
\end{figure}

Figure~\ref{fig:adaptive-control-tracking} shows that the control strategies
induce markedly different training dynamics. After the initial 500 updates,
the fine-grained controller keeps the moving-average recovery accuracy within
the target interval for $96.9\%$ of the windows. The corresponding fractions
are only $12.4\%$ for coarse-grained control and $14.6\%$ for the fixed-ratio
baseline. Both alternatives drift above the upper boundary and eventually
remain around $0.82$--$0.86$, indicating that the injected prefixes no longer
increase in difficulty quickly enough to compensate for the improving policy.
In contrast, per-problem adaptation continuously rebalances difficulty and
keeps the aggregate recovery signal close to the intended operating point.

\begin{figure}[t]
\centering
\includegraphics[width=0.5\linewidth]{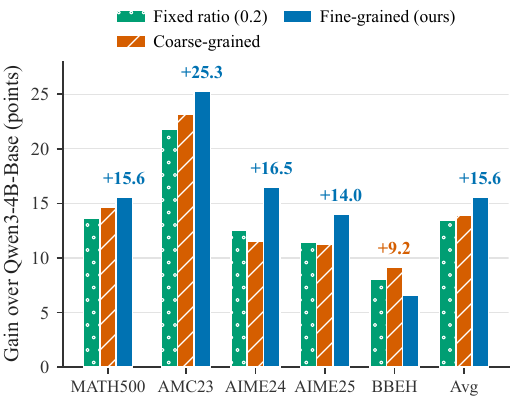}
\caption{
Absolute benchmark-accuracy gains over Qwen3-4B-Base for fixed, coarse-grained, and fine-grained noise-intensity control. 
The average is computed over the five displayed benchmarks.
}
\label{fig:adaptive-control-performance}
\end{figure}

The improved target tracking also corresponds to stronger downstream reasoning
performance (Figure~\ref{fig:adaptive-control-performance}). Fine-grained
control achieves an average gain of $15.6$ points over Qwen3-4B-Base, compared
with $13.9$ points for coarse-grained control and $13.5$ points for the fixed
ratio. It gives the largest improvement on four of the five benchmarks, with
the clearest advantage on AIME24 ($+16.5$ points versus $+11.5$ and $+12.5$
for coarse and fixed control). Coarse-grained control performs best on BBEH,
but its advantage there does not offset the larger gains of fine-grained
control on the other benchmarks. Together, these results indicate that
per-problem noise adaptation not only regulates training difficulty more
faithfully, but also yields the strongest overall reasoning performance in
this ablation.

\subsection{Generalization to Agentic Decision-Making}
To evaluate whether DenoiseRL generalizes beyond mathematical reasoning to a different domain, we apply it to ALFWorld~\cite{shridhar2021alfworldaligningtextembodied}, a text-based environment for multi-step household decision-making.
We compare the base model, standard GRPO, and DenoiseRL-GRPO on both the seen and unseen evaluation splits.

Since ALFWorld requires multi-step interaction, we adapt the line-level prefix control used for mathematical reasoning to step-level prefix control: prefix length is measured and updated in environment interaction steps rather than lines.
Analogous to using the 7,500 problems in MATH-7.5K as the units of prefix control, we maintain an independent DenoiseRL prefix-length curriculum for each of the six ALFWorld task categories.
We provide the detailed training settings in Appendix~D of the supplementary material.
We use Qwen2.5-7B-Instruct as the policy model and Qwen2.5-1.5B-Instruct~\cite{qwen2} as the weak model that supplies noisy action prefixes.

\begin{table*}[t]
\centering
\caption{
Task-wise success rates (\%) on the seen and unseen evaluation splits of ALFWorld.
For brevity, Pick, Look, Clean, Heat, Cool, and Pick2 abbreviate Pick \& Place, Examine in Light, Clean \& Place, Heat \& Place, Cool \& Place, and Pick Two \& Place, respectively; All reports overall performance.
The best result in each column within each split is shown in \textbf{bold}.
}
\label{tab:alfworld}
\small
\setlength{\tabcolsep}{5pt}
\renewcommand{\arraystretch}{1.08}
\begin{tabular*}{0.92\textwidth}{@{\extracolsep{\fill}}lccccccc@{}}
\toprule
\textbf{Method}
& \textbf{Pick}
& \textbf{Look}
& \textbf{Clean}
& \textbf{Heat}
& \textbf{Cool}
& \textbf{Pick2}
& \textbf{All} \\
\midrule
\multicolumn{8}{c}{\textit{Seen}} \\
\midrule
Base Model     & 25.7 & 38.5 & 7.4 & 6.3 & 4.0 & 5.2 & 13.6 \\
GRPO           & 91.4 & 61.5 & 92.6 & 81.2 & 72.0 & 70.8 & 80.7 \\
\textbf{DenoiseRL-GRPO} & \textbf{100.0} & \textbf{92.3} & \textbf{100.0} & \textbf{93.8} & \textbf{92.0} & \textbf{95.8} & \textbf{96.3} \\
\midrule
\multicolumn{8}{c}{\textit{Unseen}} \\
\midrule
Base Model     & 25.0 & 27.8 & 9.7 & 4.3 & 9.5 & 0.0 & 12.7 \\
GRPO           & 79.2 & 94.4 & 80.6 & 82.6 & \textbf{85.7} & 52.9 & 79.9 \\
\textbf{DenoiseRL-GRPO} & \textbf{87.5} & \textbf{100.0} & \textbf{90.3} & \textbf{87.0} & \textbf{85.7} & \textbf{76.5} & \textbf{88.1} \\
\bottomrule
\end{tabular*}
\end{table*}

\begin{table*}[t]
\centering
\caption{A representative recovery case. Red text marks key errors in the noisy prefix, while bold text highlights recovery behavior in the model continuation.}
\label{tab:case-study}
\footnotesize
\setlength{\tabcolsep}{5pt}
\renewcommand{\arraystretch}{1.02}
\begin{tabular}{p{0.14\linewidth}p{0.80\linewidth}}
\toprule
\textbf{Part} & \textbf{Content} \\
\midrule

Problem &
A school has between $150$ and $200$ students enrolled. 
If one student is absent, the students can be divided equally into six sections. 
What is the sum of all possible enrollment numbers? \\

\midrule

Noisy Prefix &
Let the number of students be $n$. Since one student is absent, $n-1$ is divisible by $6$, so
$n-1=6k$ and $n=6k+1$.
\textit{[omitted: several attempted substitutions for $k$]}

For $k=17$, since $6\times17+1=103$, 
$103$ is between $150$ and $200$. 
For $k=18$, since $6\times18+1=109$, 
$109$ is between $150$ and $200$. 
\textit{[omitted: more invalid candidates below $150$]}

Therefore, the possible values for $n$ are $193,199$, and the sum is
\textcolor{red}{$193+199=392$}. \\

\midrule

Model Continuation &
\textbf{Let's re-evaluate the problem step by step} to ensure we find all possible values of $n$ that satisfy the given conditions. 
Since $n-1\equiv0\pmod 6$, we have $n\equiv1\pmod 6$ and $n=6k+1$.

Now, find all integers $k$ such that $150<n<200$:
$150<6k+1<200$, hence $149<6k<199$ and
$\frac{149}{6}<k<\frac{199}{6}$, i.e.,
$24.833\ldots<k<33.166\ldots$. 
Since $k$ must be an integer, $k$ ranges from $25$ to $33$ inclusive.

The corresponding values of $n$ are 
$151,157,163,169,175,181,187,193,199$. 
Finally, $151+157+\cdots+199=1575$, so the final answer is $\boxed{1575}$.
\\
\bottomrule
\end{tabular}
\end{table*}

As shown in Table~\ref{tab:alfworld}, DenoiseRL-GRPO achieves the best overall success rate on both evaluation splits, reaching 96.3\% on seen environments and 88.1\% on unseen environments.
This improves over standard GRPO by 15.6 and 8.2 percentage points, respectively.
DenoiseRL-GRPO performs best on every task category in the seen split and is best or tied for best on every category in the unseen split, demonstrating that recovery training transfers effectively from mathematical reasoning to multi-step agentic decision-making.

\subsection{Case Study}
\label{sec:case-study}

The purpose of this case study is to examine whether DenoiseRL induces genuine denoising and recovery behavior rather than merely encouraging the policy to continue from a noisy prefix. 
In particular, we inspect a rollout where the prefix contains a partially correct derivation but reaches an incorrect answer due to faulty enumeration. 
As shown in Table~\ref{tab:case-study}, the model continuation does not follow the erroneous conclusion. 
Instead, it re-checks the core constraint, recomputes the feasible range, and repairs the final answer.

This suggests that denoise rollouts teach the model to use weak-model errors as perturbations: the policy learns to preserve useful partial reasoning while correcting the failure modes that lead to wrong answers.
We provide more cases in Appendix~C of the supplementary
material.


\section{Conclusion}
We propose DenoiseRL, a recovery-oriented RL framework that improves reasoning capability by training models to recover from incorrect intermediate trajectories generated by weak models.
Instead of relying on stronger teachers or manually curated hard datasets, DenoiseRL converts weak-model failures into structured perturbations and controls their intensity through a sample-level adaptive curriculum.
By adjusting each problem's prefix length according to online recovery accuracy, the curriculum tracks the policy's evolving capability boundary and keeps training examples informative as the model improves.
Our experiments further show that fine-grained per-problem control outperforms fixed and coarse-grained strategies, while noise from an external weak model provides more persistent late-stage interference and stronger overall performance than self-generated or random noise.
Our results also show that this recovery-centric optimization consistently improves reasoning and interactive decision-making while strengthening the model's ability to revise flawed reasoning.
Overall, DenoiseRL offers a scalable post-training perspective in which model errors provide both the learning signal and the basis for an adaptive  curriculum.

\section*{Limitations}
Although DenoiseRL demonstrates strong empirical performance, several limitations remain. 
First, the effectiveness of generated perturbations still depends on the behavior of the weak models used for corruption generation. 
If the weak models produce errors that are overly trivial, repetitive, or unrealistic, the resulting recovery signal may provide limited training value and fail to induce meaningful improvements. 

Second, although DenoiseRL is formulated as a general framework for learning from perturbed reasoning states, our experiments instantiate such perturbations only as prefixes of failed weak-model trajectories. It remains unclear whether the same recovery objective extends to other representations or corruption mechanisms, such as erroneous intermediate conclusions, structured plans, tool-use histories, or latent reasoning states. Exploring these alternatives may broaden DenoiseRL beyond settings in which reasoning states can be naturally realized as prefixes.

\appendix
\clearpage
\label{sec:appendix}
\section{Training Prompt and Hyperparameters for Mathematical Tasks}
\subsection{Training Prompt}
\label{sec:prompt}
\begin{mybox}[label=box:prompt_q]{}
\textbf{Prompt $z_{q}$:}\\
System \\
You are a helpful assistant. \\
User \\
\{question\} \\
Please reasoning step by step, and put your final answer in \textbackslash\textbackslash boxed\{\}.
\end{mybox}
\subsection{Hyperparameters}

\paragraph{Optimization and rollout.}
We train Qwen3-4B-Base with GRPO using an active batch of 16 prompts and
$K=16$ recovery rollouts per prompt, without additional clean rollouts.
The training and PPO mini-batch sizes are both 16.
We use a learning rate of $10^{-6}$ without warmup or weight decay, clip the
gradient norm at $1.0$, and set both PPO clipping bounds to $0.2$.
KL regularization is disabled in both the reward and actor loss.
The maximum prompt and response lengths are 8,192 and 4,096 tokens,
respectively.
Training rollouts use temperature $1.0$, top-p $1.0$, and unrestricted top-k;
validation uses temperature $0.6$ and top-p $0.95$.

\paragraph{DenoiseRL curriculum.}
We use line-level prefixes from incorrect weak-model trajectories and maintain
an independent prefix ratio for each problem.
The ratio is initialized and lower-bounded at $0$, upper-bounded at $0.5$, and
updated toward target recovery accuracy $0.75$ with step size $\alpha=0.2$.
The controller uses a five-observation history, requires at least two
observations before assessing stability, and marks a problem stable when the
absolute recent ratio slope is at most $0.02$.
Replacement problems inherit the post-update mean ratio of the active batch.
We enable the dynamic length reward with minimum factor $0.9$ for all recovery
rollouts and apply no additional response-clipping penalty.

\section{Training Efficiency}
\label{sec:training-efficiency}

\paragraph{Offline noisy-prefix construction.}
For mathematical reasoning, DenoiseRL samples incorrect trajectories from the
Qwen2.5-1.5B-Instruct weak model once and stores the resulting prefixes before
policy optimization.
This preprocessing is lightweight because it uses a small model, is performed
only once, and can be reused across training runs.
More importantly, no weak-model inference is required inside the RL loop:
during training, DenoiseRL only retrieves a stored prefix and selects the
portion specified by the current per-problem prefix ratio.
Prefix construction therefore consumes none of the online policy-training time
reported below.

\begin{figure}[t]
    \centering
    \includegraphics[width=0.5\columnwidth]{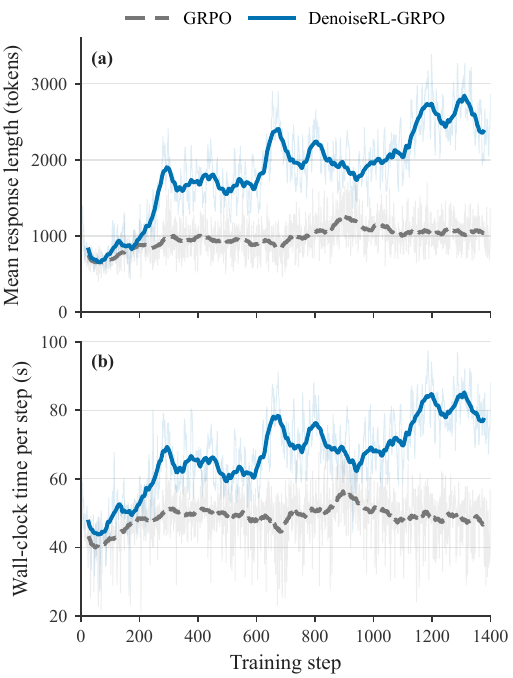}
    \caption{Response length and wall-clock training time on Qwen3-4B-Base
    under matched batch and rollout settings ($B=16$, $K=16$).
    Faint traces show per-step observations and curves show 50-step moving
    averages over the shared first 1,400 updates.
    DenoiseRL's noisy prefixes are sampled offline and their construction is not
    included in the reported training time.}
    \label{fig:training-efficiency}
\end{figure}

\paragraph{Online training time.}
Figure~\ref{fig:training-efficiency} shows that the wall-clock difference from
standard GRPO closely follows the length of the policy-generated responses.
Across the shared 1,400-step horizon, DenoiseRL-GRPO produces 1,843 tokens per
response on average, compared with 974 for GRPO, a $1.89\times$ increase.
Its corresponding mean step time is 67.0 seconds, compared with 48.8 seconds
for GRPO, a smaller $1.37\times$ increase.
The per-step Pearson correlation between response length and step time is
$0.96$ for DenoiseRL-GRPO and $0.84$ for GRPO.
The two panels also exhibit the same late-stage rises and local peaks.
These results indicate that the observed runtime gap is primarily associated
with the longer on-policy generations induced by recovery training, rather
than with online noisy-prefix acquisition.
The longer responses are also consistent with DenoiseRL eliciting a stronger
test-time-scaling behavior in which the policy allocates more tokens to
reasoning.

\section{Supplementary Cases}
\label{sec:cases}

We provide two additional recovery cases to further illustrate the behavior induced by DenoiseRL.
For readability, we omit some repetitive intermediate calculations while preserving the key prefix-continuation boundary determined by the recorded token-level prefix length.
Red text marks the error or failure mode in the noisy prefix, while bold text highlights recovery behavior in the model continuation.

\begin{table*}[t]
\centering
\caption{A supplementary recovery case on probability reasoning. The noisy prefix contains an incorrect local probability statement, while the continuation switches to a correct counting formulation.}
\label{tab:supp-case-prob}
\footnotesize
\setlength{\tabcolsep}{5pt}
\renewcommand{\arraystretch}{1.02}
\begin{tabular}{p{0.14\linewidth}p{0.80\linewidth}}
\toprule
\textbf{Part} & \textbf{Content} \\
\midrule

Problem &
The reality game show \textit{Survivor} is played with $16$ people divided into two tribes of $8$.
In the first episode, two people get homesick and quit.
If every person has an equal chance of being one of the two quitters, what is the probability that both people who quit are from the same tribe? \\

\midrule

Noisy Prefix &
To determine the probability that both people who quit are from the same tribe, we can break the problem down into manageable steps.

There are $16$ people in the game, divided equally into two tribes of $8$.
For Tribe A, there are $8$ people.
The prefix then states that
\textcolor{red}{the probability that a specific person quits is $\frac{1}{15}$},
because one person is chosen out of $16$ and the remaining choices are considered.
\textit{[omitted: the prefix begins from a local individual-quitter calculation rather than the correct pair-level sample space.]} \\

\midrule

Model Continuation &
\textbf{Since we are finding the probability of both quitters being from the same tribe, we need a different approach.}
The continuation constructs the correct sample space:
\[
\binom{16}{2}=120.
\]
The favorable cases are those where both quitters come from Tribe A or both come from Tribe B:
\[
\binom{8}{2}+\binom{8}{2}=28+28=56.
\]
Therefore,
\[
\frac{56}{120}=\frac{7}{15},
\]
so the final answer is
\[
\boxed{\boldsymbol{\frac{7}{15}}}.
\] \\

\bottomrule
\end{tabular}
\end{table*}

The first case in Table~\ref{tab:supp-case-prob} shows recovery from a misleading local probability calculation.
The prefix contains useful structural information that there are two tribes of size $8$, but it frames the problem around the probability of an individual quitter, which is not the most direct sample space.
The continuation does not elaborate on this flawed direction.
Instead, it explicitly changes strategy, counts unordered pairs of quitters, and compares favorable same-tribe pairs against all possible pairs.
This behavior indicates that the policy can preserve useful problem facts from the prefix while rejecting an unproductive probabilistic framing.

\begin{table*}[t]
\centering
\caption{A supplementary recovery case on algebraic reasoning. The noisy prefix attempts explicit root search, while the continuation recovers by using Vieta's formulas.}
\label{tab:supp-case-vieta}
\footnotesize
\setlength{\tabcolsep}{5pt}
\renewcommand{\arraystretch}{1.02}
\begin{tabular}{p{0.14\linewidth}p{0.80\linewidth}}
\toprule
\textbf{Part} & \textbf{Content} \\
\midrule

Problem &
The graphs of $y=x^3-3x+2$ and $x+4y=4$ intersect in the points
$(x_1,y_1)$, $(x_2,y_2)$, and $(x_3,y_3)$.
If $x_1+x_2+x_3=A$ and $y_1+y_2+y_3=B$, compute the ordered pair $(A,B)$. \\

\midrule

Noisy Prefix &
From $x+4y=4$, the prefix derives
\[
y=1-\frac{x}{4}.
\]
Substituting this into $y=x^3-3x+2$ gives
\[
1-\frac{x}{4}=x^3-3x+2,
\]
and hence
\[
4x^3-11x+4=0.
\]
The prefix then attempts to find the individual roots using the Rational Root Theorem, testing candidates such as
$x=1$, $x=\frac12$, $x=-\frac12$, $x=2$, and $x=-1$.
\textit{[omitted: several direct substitutions showing that these candidates are not roots.]}

\textcolor{red}{The prefix keeps searching for explicit roots, although the problem only asks for the sums of the coordinates.} \\

\midrule

Model Continuation &
\textbf{It seems the rational root theorem isn't yielding roots easily. Let's try another method.}
Given
\[
4x^3-11x+4=0,
\]
the continuation writes it as
\[
4x^3+0x^2-11x+4=0.
\]
By Vieta's formulas,
\[
x_1+x_2+x_3=-\frac{0}{4}=0,
\]
so $A=0$.
Using $y=1-\frac{x}{4}$,
\[
y_1+y_2+y_3
=
\left(1-\frac{x_1}{4}\right)
+
\left(1-\frac{x_2}{4}\right)
+
\left(1-\frac{x_3}{4}\right)
=
3-\frac{x_1+x_2+x_3}{4}
=
3.
\]
Thus, the final answer is
\[
\boxed{\boldsymbol{(0,3)}}.
\] \\

\bottomrule
\end{tabular}
\end{table*}

The second case in Table~\ref{tab:supp-case-vieta} illustrates recovery from an inefficient search procedure.
The prefix correctly transforms the intersection problem into a cubic equation, but then pursues explicit roots through repeated rational-root checks.
This is unnecessary for the target quantity, since the question asks only for aggregate coordinate sums.
The continuation identifies the dead end, switches to Vieta's formulas, and obtains $A$ directly from the missing $x^2$ coefficient.
It then transfers this aggregate information back through the linear relation $y=1-\frac{x}{4}$ to compute $B$.
Together with the main-text case, these examples suggest that denoise rollouts do more than encourage continuation from noisy contexts: they train the model to re-assess the current reasoning state, retain useful intermediate structure, and replace flawed or inefficient local strategies with solution-relevant ones.

\section{Hyperparameters for Agentic Tasks}
\label{sec:agentic-hyperparameters}

\paragraph{Model, environment, and rollout.}
We use Qwen2.5-7B-Instruct as the policy and Qwen2.5-1.5B-Instruct as the online
weak model.
Each optimizer step uses a batch of 16 ALFWorld tasks and $K=16$ recovery
rollouts per task, with no separate standard rollout.
The weak model generates one prefix candidate per task group using the
\texttt{full\_then\_ratio} strategy.
Episodes are capped at 50 environment steps, and the maximum prompt and
response lengths are 4,096 and 512 tokens, respectively.
Training generation uses temperature $1.0$ and top-p $1.0$; validation uses
temperature $0.6$, top-p $0.95$, and one sample per environment.

\paragraph{Optimization and curriculum.}
We optimize with GRPO for up to 500 epochs using a learning rate of $10^{-6}$,
a PPO mini-batch size of 256, and a per-GPU micro-batch size of 32.
KL is excluded from the reward, while the actor uses a low-variance KL loss
with coefficient $0.01$.
Invalid actions receive a penalty of $0.1$.
For the step-level DenoiseRL curriculum, all game files belonging to the same
ALFWorld task category share one dynamic prefix ratio, while fresh game files
are sampled at every optimizer step.
We use $\rho_{\mathrm{init}}=\rho_{\min}=0$, $\rho_{\max}=0.5$, target recovery
accuracy $0.75$, update step size $\alpha=0.2$, and shuffle seed $0$.

\clearpage

\bibliographystyle{plainnat}
\bibliography{main}

@article{burns2023weak,
  title={Weak-to-strong generalization: Eliciting strong capabilities with weak supervision},
  author={Burns, Collin and Izmailov, Pavel and Kirchner, Jan Hendrik and Baker, Bowen and Gao, Leo and Aschenbrenner, Leopold and Chen, Yining and Ecoffet, Adrien and Joglekar, Manas and Leike, Jan and others},
  journal={arXiv preprint arXiv:2312.09390},
  year={2023}
}

@inproceedings{huang2024large,
  title={Large language models cannot self-correct reasoning yet},
  author={Huang, Jie and Chen, Xinyun and Mishra, Swaroop and Zheng, Huaixiu Steven and Yu, Adams and Song, Xinying and Zhou, Denny},
  booktitle={International conference on learning representations},
  volume={2024},
  pages={32808--32824},
  year={2024}
}

@article{ouyang2022training,
  title={Training language models to follow instructions with human feedback},
  author={Ouyang, Long and Wu, Jeffrey and Jiang, Xu and Almeida, Diogo and Wainwright, Carroll and Mishkin, Pamela and Zhang, Chong and Agarwal, Sandhini and Slama, Katarina and Ray, Alex and others},
  journal={Advances in neural information processing systems},
  volume={35},
  pages={27730--27744},
  year={2022}
}

@article{shao2024deepseekmath,
  title={Deepseekmath: Pushing the limits of mathematical reasoning in open language models},
  author={Shao, Zhihong and Wang, Peiyi and Zhu, Qihao and Xu, Runxin and Song, Junxiao and Bi, Xiao and Zhang, Haowei and Zhang, Mingchuan and Li, YK and Wu, Yang and others},
  journal={arXiv preprint arXiv:2402.03300},
  year={2024}
}

@article{yu2026dapo,
  title={Dapo: An open-source llm reinforcement learning system at scale},
  author={Yu, Qiying and Zhang, Zheng and Zhu, Ruofei and Yuan, Yufeng and Zuo, Xiaochen and Yue, Yu and Dai, Weinan and Fan, Tiantian and Liu, Gaohong and Liu, Lingjun and others},
  journal={Advances in Neural Information Processing Systems},
  volume={38},
  pages={113222--113244},
  year={2026}
}

@inproceedings{lightman2024let,
  title={Let's verify step by step},
  author={Lightman, Hunter and Kosaraju, Vineet and Burda, Yuri and Edwards, Harrison and Baker, Bowen and Lee, Teddy and Leike, Jan and Schulman, John and Sutskever, Ilya and Cobbe, Karl},
  booktitle={International Conference on Learning Representations},
  volume={2024},
  pages={39578--39601},
  year={2024}
}

@article{khalifa2025process,
  title={Process reward models that think},
  author={Khalifa, Muhammad and Agarwal, Rishabh and Logeswaran, Lajanugen and Kim, Jaekyeom and Peng, Hao and Lee, Moontae and Lee, Honglak and Wang, Lu},
  journal={arXiv preprint arXiv:2504.16828},
  year={2025}
}

@article{kim2025training,
  title={Training-free Uncertainty Guidance for Complex Visual Tasks with MLLMs},
  author={Kim, Sanghwan and Xiao, Rui and Alaniz, Stephan and Xian, Yongqin and Akata, Zeynep},
  journal={arXiv preprint arXiv:2510.00705},
  year={2025}
}

@article{cai2025training,
  title={Training-free group relative policy optimization},
  author={Cai, Yuzheng and Cai, Siqi and Shi, Yuchen and Xu, Zihan and Chen, Lichao and Qin, Yulei and Tan, Xiaoyu and Li, Gang and Li, Zongyi and Lin, Haojia and others},
  journal={arXiv preprint arXiv:2510.08191},
  year={2025}
}

@inproceedings{yao2025revisiting,
  title={Revisiting weak-to-strong generalization in theory and practice: Reverse KL vs. forward KL},
  author={Yao, Wei and Yang, Wenkai and Wang, Ziqiao and Lin, Yankai and Liu, Yong},
  booktitle={Findings of the Association for Computational Linguistics: ACL 2025},
  pages={2860--2888},
  year={2025}
}

@article{qwen2,
      title={Qwen2 Technical Report}, 
      author={An Yang and Baosong Yang and Binyuan Hui and Bo Zheng and Bowen Yu and Chang Zhou and Chengpeng Li and Chengyuan Li and Dayiheng Liu and Fei Huang and Guanting Dong and Haoran Wei and Huan Lin and Jialong Tang and Jialin Wang and Jian Yang and Jianhong Tu and Jianwei Zhang and Jianxin Ma and Jin Xu and Jingren Zhou and Jinze Bai and Jinzheng He and Junyang Lin and Kai Dang and Keming Lu and Keqin Chen and Kexin Yang and Mei Li and Mingfeng Xue and Na Ni and Pei Zhang and Peng Wang and Ru Peng and Rui Men and Ruize Gao and Runji Lin and Shijie Wang and Shuai Bai and Sinan Tan and Tianhang Zhu and Tianhao Li and Tianyu Liu and Wenbin Ge and Xiaodong Deng and Xiaohuan Zhou and Xingzhang Ren and Xinyu Zhang and Xipin Wei and Xuancheng Ren and Yang Fan and Yang Yao and Yichang Zhang and Yu Wan and Yunfei Chu and Yuqiong Liu and Zeyu Cui and Zhenru Zhang and Zhihao Fan},
      journal={arXiv preprint arXiv:2407.10671},
      year={2024}
}

@misc{qwen3technicalreport,
      title={Qwen3 Technical Report}, 
      author={Qwen Team},
      year={2025},
      eprint={2505.09388},
      archivePrefix={arXiv},
      primaryClass={cs.CL},
      url={https://arxiv.org/abs/2505.09388}, 
}

@article{hendrycks2021measuring,
  title={Measuring mathematical problem solving with the math dataset},
  author={Hendrycks, Dan and Burns, Collin and Kadavath, Saurav and Arora, Akul and Basart, Steven and Tang, Eric and Song, Dawn and Steinhardt, Jacob},
  journal={arXiv preprint arXiv:2103.03874},
  year={2021}
}

@online{aops_amc,
  title     = {AMC Problems and Solutions},
  author    = {{Art of Problem Solving}},
  year      = {2025},
  url       = {https://artofproblemsolving.com/wiki/index.php?title=AMC_Problems_and_Solutions}
}

@online{aops_aime,
  title     = {AIME Problems and Solutions},
  author    = {{Art of Problem Solving}},
  year      = {2025},
  url       = {https://artofproblemsolving.com/wiki/index.php/AIME_Problems_and_Solutions}
}

@article{schulman2017proximal,
  title={Proximal policy optimization algorithms},
  author={Schulman, John and Wolski, Filip and Dhariwal, Prafulla and Radford, Alec and Klimov, Oleg},
  journal={arXiv preprint arXiv:1707.06347},
  year={2017}
}

@inproceedings{suzgun2023challenging,
  title={Challenging big-bench tasks and whether chain-of-thought can solve them},
  author={Suzgun, Mirac and Scales, Nathan and Sch{\"a}rli, Nathanael and Gehrmann, Sebastian and Tay, Yi and Chung, Hyung Won and Chowdhery, Aakanksha and Le, Quoc and Chi, Ed and Zhou, Denny and others},
  booktitle={Findings of the Association for Computational Linguistics: ACL 2023},
  pages={13003--13051},
  year={2023}
}

@misc{lei2026stepbackprefiximportance,
      title={A Step Back: Prefix Importance Ratio Stabilizes Policy Optimization}, 
      author={Shiye Lei and Zhihao Cheng and Dacheng Tao},
      year={2026},
      eprint={2601.22718},
      archivePrefix={arXiv},
      primaryClass={cs.AI},
      url={https://arxiv.org/abs/2601.22718}, 
}

@article{hu2025reinforce++,
  title={Reinforce++: A simple and efficient approach for aligning large language models},
  author={Hu, Jian},
  journal={arXiv e-prints},
  pages={arXiv--2501},
  year={2025}
}

@misc{ahmadian2024basicsrevisitingreinforcestyle,
      title={Back to Basics: Revisiting REINFORCE Style Optimization for Learning from Human Feedback in LLMs}, 
      author={Arash Ahmadian and Chris Cremer and Matthias Gallé and Marzieh Fadaee and Julia Kreutzer and Olivier Pietquin and Ahmet Üstün and Sara Hooker},
      year={2024},
      eprint={2402.14740},
      archivePrefix={arXiv},
      primaryClass={cs.LG},
      url={https://arxiv.org/abs/2402.14740}, 
}

@misc{xiao2025bnpobetanormalizationpolicy,
      title={BNPO: Beta Normalization Policy Optimization}, 
      author={Changyi Xiao and Mengdi Zhang and Yixin Cao},
      year={2025},
      eprint={2506.02864},
      archivePrefix={arXiv},
      primaryClass={cs.LG},
      url={https://arxiv.org/abs/2506.02864}, 
}

@misc{zhang2026reinforcementawareknowledgedistillationllm,
      title={Reinforcement-aware Knowledge Distillation for LLM Reasoning}, 
      author={Zhaoyang Zhang and Shuli Jiang and Yantao Shen and Yuting Zhang and Dhananjay Ram and Shuo Yang and Zhuowen Tu and Wei Xia and Stefano Soatto},
      year={2026},
      eprint={2602.22495},
      archivePrefix={arXiv},
      primaryClass={cs.LG},
      url={https://arxiv.org/abs/2602.22495}, 
}

@misc{yu2026knowrlboostingllmreasoning,
      title={KnowRL: Boosting LLM Reasoning via Reinforcement Learning with Minimal-Sufficient Knowledge Guidance}, 
      author={Linhao Yu and Tianmeng Yang and Siyu Ding and Renren Jin and Naibin Gu and Xiangzhao Hao and Shuaiyi Nie and Deyi Xiong and Weichong Yin and Yu Sun and Hua Wu},
      year={2026},
      eprint={2604.12627},
      archivePrefix={arXiv},
      primaryClass={cs.AI},
      url={https://arxiv.org/abs/2604.12627}, 
}

@misc{he2025deepmath103klargescalechallengingdecontaminated,
      title={DeepMath-103K: A Large-Scale, Challenging, Decontaminated, and Verifiable Mathematical Dataset for Advancing Reasoning}, 
      author={Zhiwei He and Tian Liang and Jiahao Xu and Qiuzhi Liu and Xingyu Chen and Yue Wang and Linfeng Song and Dian Yu and Zhenwen Liang and Wenxuan Wang and Zhuosheng Zhang and Rui Wang and Zhaopeng Tu and Haitao Mi and Dong Yu},
      year={2025},
      eprint={2504.11456},
      archivePrefix={arXiv},
      primaryClass={cs.CL},
      url={https://arxiv.org/abs/2504.11456}, 
}

@misc{yuan2026incentivizingstrongreasoningweak,
      title={Incentivizing Strong Reasoning from Weak Supervision}, 
      author={Yige Yuan and Teng Xiao and Shuchang Tao and Xue Wang and Jinyang Gao and Bolin Ding and Bingbing Xu},
      year={2026},
      eprint={2505.20072},
      archivePrefix={arXiv},
      primaryClass={cs.CL},
      url={https://arxiv.org/abs/2505.20072}, 
}

@misc{lang2025selectiveweaktostronggeneralization,
      title={Selective Weak-to-Strong Generalization}, 
      author={Hao Lang and Fei Huang and Yongbin Li},
      year={2025},
      eprint={2511.14166},
      archivePrefix={arXiv},
      primaryClass={cs.CL},
      url={https://arxiv.org/abs/2511.14166}, 
}

@inproceedings{liu-etal-2025-atlantis,
    title = "{ATLANTIS}: Weak-to-Strong Learning via Importance Sampling",
    author = "Liu, Yi  and
      Wang, Guoyin  and
      Li, Shicheng  and
      Song, Feifan  and
      Sun, Xu",
    editor = "Che, Wanxiang  and
      Nabende, Joyce  and
      Shutova, Ekaterina  and
      Pilehvar, Mohammad Taher",
    booktitle = "Proceedings of the 63rd Annual Meeting of the Association for Computational Linguistics (Volume 1: Long Papers)",
    month = jul,
    year = "2025",
    address = "Vienna, Austria",
    publisher = "Association for Computational Linguistics",
    url = "https://aclanthology.org/2025.acl-long.52/",
    doi = "10.18653/v1/2025.acl-long.52",
    pages = "1042--1052",
    ISBN = "979-8-89176-251-0"
}

@article{xu2026scaler,
  title={SCALER: Synthetic Scalable Adaptive Learning Environment for Reasoning},
  author={Xu, Caijun and Xiao, Changyi and Peng, Zhongyuan and Wang, Xinrun and Cao, Yixin},
  journal={arXiv preprint arXiv:2601.04809},
  year={2026}
}

@misc{zhan2026mathsmithextremelyhardmathematical,
      title={MathSmith: Towards Extremely Hard Mathematical Reasoning by Forging Synthetic Problems with a Reinforced Policy}, 
      author={Shaoxiong Zhan and Yanlin Lai and Ziyu Lu and Dahua Lin and Ziqing Yang and Fei Tan},
      year={2026},
      eprint={2508.05592},
      archivePrefix={arXiv},
      primaryClass={cs.CL},
      url={https://arxiv.org/abs/2508.05592}, 
}

@misc{liu2025synlogicsynthesizingverifiablereasoning,
      title={SynLogic: Synthesizing Verifiable Reasoning Data at Scale for Learning Logical Reasoning and Beyond}, 
      author={Junteng Liu and Yuanxiang Fan and Zhuo Jiang and Han Ding and Yongyi Hu and Chi Zhang and Yiqi Shi and Shitong Weng and Aili Chen and Shiqi Chen and Yunan Huang and Mozhi Zhang and Pengyu Zhao and Junjie Yan and Junxian He},
      year={2025},
      eprint={2505.19641},
      archivePrefix={arXiv},
      primaryClass={cs.AI},
      url={https://arxiv.org/abs/2505.19641}, 
}

@article{yan2026learning,
  title={Learning to reason under off-policy guidance},
  author={Yan, Jianhao and Li, Yafu and Hu, Zican and Wang, Zhi and Cui, Ganqu and Qu, Xiaoye and Cheng, Yu and Zhang, Yue},
  journal={Advances in Neural Information Processing Systems},
  volume={38},
  pages={117157--117186},
  year={2026}
}

@article{setlur2026reuse,
  title={Reuse your FLOPs: Scaling RL on Hard Problems by Conditioning on Very Off-Policy Prefixes},
  author={Setlur, Amrith and Wang, Zijian and Cohen, Andrew and Rashidinejad, Paria and Xie, Sang Michael},
  journal={arXiv preprint arXiv:2601.18795},
  year={2026}
}

@article{qu2026pope,
  title={POPE: Learning to Reason on Hard Problems via Privileged On-Policy Exploration},
  author={Qu, Yuxiao and Setlur, Amrith and Smith, Virginia and Salakhutdinov, Ruslan and Kumar, Aviral},
  journal={arXiv preprint arXiv:2601.18779},
  year={2026}
}

@inproceedings{sun2026well,
  title={Well Begun, Half Done: Reinforcement Learning with Prefix Optimization for LLM Reasoning},
  author={Sun, Yiliu and Zhao, Zicheng and Wei, Yang and Zhang, Yanfang and Gong, Chen},
  booktitle={Proceedings of the AAAI Conference on Artificial Intelligence},
  volume={40},
  number={39},
  pages={33144--33152},
  year={2026}
}

@article{guo2025deepseek,
  title={Deepseek-r1: Incentivizing reasoning capability in llms via reinforcement learning},
  author={Guo, Daya and Yang, Dejian and Zhang, Haowei and Song, Junxiao and Wang, Peiyi and Zhu, Qihao and Xu, Runxin and Zhang, Ruoyu and Ma, Shirong and Bi, Xiao and others},
  journal={arXiv preprint arXiv:2501.12948},
  year={2025}
}

@article{liu2025evocot,
  title={Evocot: Overcoming the exploration bottleneck in reinforcement learning},
  author={Liu, Huanyu and Li, Jia and Dong, Yihong and Yu, Chang and Chen, Taozhi and Wang, Lecheng and Tao, Yongding and Gu, Bin and Li, Ge},
  journal={arXiv preprint arXiv:2508.07809},
  year={2025}
}

@inproceedings{somerstep2024statistical,
  title={A statistical framework for weak-to-strong generalization},
  author={Somerstep, Seamus and Polo, Felipe Maia and Banerjee, Moulinath and Ritov, Yaacov and Yurochkin, Mikhail and Sun, Yuekai},
  booktitle={ICML 2024 Next Generation of AI Safety Workshop},
  year={2024}
}

@article{geng2026weak,
  title={Weak-to-Strong Generalization is Nearly Inevitable (in Linear Models)},
  author={Geng, Scott and Hansen, Dutch and Li, Jerry},
  journal={arXiv preprint arXiv:2605.05742},
  year={2026}
}

@article{welleck2022generating,
  title={Generating sequences by learning to self-correct},
  author={Welleck, Sean and Lu, Ximing and West, Peter and Brahman, Faeze and Shen, Tianxiao and Khashabi, Daniel and Choi, Yejin},
  journal={arXiv preprint arXiv:2211.00053},
  year={2022}
}

@inproceedings{vincent2008extracting,
  title={Extracting and composing robust features with denoising autoencoders},
  author={Vincent, Pascal and Larochelle, Hugo and Bengio, Yoshua and Manzagol, Pierre-Antoine},
  booktitle={Proceedings of the 25th international conference on Machine learning},
  pages={1096--1103},
  year={2008}
}

@inproceedings{lewis2020bart,
  title={BART: Denoising sequence-to-sequence pre-training for natural language generation, translation, and comprehension},
  author={Lewis, Mike and Liu, Yinhan and Goyal, Naman and Ghazvininejad, Marjan and Mohamed, Abdelrahman and Levy, Omer and Stoyanov, Veselin and Zettlemoyer, Luke},
  booktitle={Proceedings of the 58th annual meeting of the association for computational linguistics},
  pages={7871--7880},
  year={2020}
}

@misc{chen2025llmssignaltheyreright,
      title={Do LLMs Signal When They're Right? Evidence from Neuron Agreement}, 
      author={Kang Chen and Yaoning Wang and Kai Xiong and Zhuoka Feng and Wenhe Sun and Haotian Chen and Yixin Cao},
      year={2025},
      eprint={2510.26277},
      archivePrefix={arXiv},
      primaryClass={cs.CL},
      url={https://arxiv.org/abs/2510.26277}, 
}

@misc{shridhar2021alfworldaligningtextembodied,
      title={ALFWorld: Aligning Text and Embodied Environments for Interactive Learning}, 
      author={Mohit Shridhar and Xingdi Yuan and Marc-Alexandre Côté and Yonatan Bisk and Adam Trischler and Matthew Hausknecht},
      year={2021},
      eprint={2010.03768},
      archivePrefix={arXiv},
      primaryClass={cs.CL},
      url={https://arxiv.org/abs/2010.03768}, 
}

@article{zhang2025critique,
  title={Critique-grpo: Advancing llm reasoning with natural language and numerical feedback},
  author={Zhang, Xiaoying and Zhang, Yipeng and Sun, Hao and Feng, Kaituo and Lu, Chaochao and Yang, Chao and Meng, Helen},
  journal={arXiv preprint arXiv:2506.03106},
  year={2025}
}

@misc{shi2023largelanguagemodelseasily,
      title={Large Language Models Can Be Easily Distracted by Irrelevant Context}, 
      author={Freda Shi and Xinyun Chen and Kanishka Misra and Nathan Scales and David Dohan and Ed Chi and Nathanael Schärli and Denny Zhou},
      year={2023},
      eprint={2302.00093},
      archivePrefix={arXiv},
      primaryClass={cs.CL},
      url={https://arxiv.org/abs/2302.00093}, 
}

@misc{anantheswaran2025cuttingnoiseboostingllm,
      title={Cutting Through the Noise: Boosting LLM Performance on Math Word Problems}, 
      author={Ujjwala Anantheswaran and Himanshu Gupta and Kevin Scaria and Shreyas Verma and Chitta Baral and Swaroop Mishra},
      year={2025},
      eprint={2406.15444},
      archivePrefix={arXiv},
      primaryClass={cs.CL},
      url={https://arxiv.org/abs/2406.15444}, 
}

@misc{parashar2026curriculumreinforcementlearningeasy,
      title={Curriculum Reinforcement Learning from Easy to Hard Tasks Improves LLM Reasoning}, 
      author={Shubham Parashar and Shurui Gui and Xiner Li and Hongyi Ling and Sushil Vemuri and Blake Olson and Eric Li and Yu Zhang and James Caverlee and Dileep Kalathil and Shuiwang Ji},
      year={2026},
      eprint={2506.06632},
      archivePrefix={arXiv},
      primaryClass={cs.LG},
      url={https://arxiv.org/abs/2506.06632}, 
}

\end{document}